\documentclass{article}
\usepackage{spconf,amsmath,graphicx}
\usepackage{xcolor}
\usepackage{hyperref}
\usepackage{enumitem}
\usepackage{caption}
\newlength{\bibitemsep}
\newlength{\bibparskip}
\let\oldthebibliography\thebibliography
\renewcommand\thebibliography[1]{%
  \small
  \oldthebibliography{#1}%
  \setlength{\parskip}{\bibitemsep}%
  \setlength{\itemsep}{\bibparskip}%
}
\let\Footnotesize=\footnotesize  
\let\footnotesize=\scriptsize
\usepackage{enumitem}
\setlist{nosep, leftmargin=14pt}

\usepackage{mwe} 

\title{Video-based Surgical Tool-tip and Keypoint Tracking using Multi-frame Context-driven Deep Learning Models}
\name{Bhargav Ghanekar, Lianne R. Johnson, Jacob L. Laughlin, Marcia K. O'Malley, Ashok Veeraraghavan}
\address{Rice University, Houston TX USA}
\begin{document}
%
\maketitle
\begin{abstract}
    Automated tracking of surgical tool keypoints in robotic surgery videos is an essential task for various downstream use cases such as skill assessment, expertise assessment, and the delineation of safety zones. In recent years, the explosion of deep learning for vision applications has led to many works in surgical instrument segmentation, while lesser focus has been on tracking specific tool keypoints, such as tool tips. In this work, we propose a novel, multi-frame context-driven deep learning framework to localize and track tool keypoints in surgical videos. We train and test our models on the annotated frames from the 2015 EndoVis Challenge dataset, resulting in state-of-the-art performance. By leveraging sophisticated deep learning models and multi-frame context, we achieve 90\% keypoint detection accuracy and a localization RMS error of 5.27 pixels. Results on a self-annotated JIGSAWS dataset with more challenging scenarios also show that the proposed multi-frame models can accurately track tool-tip and tool-base keypoints, with ${<}4.2$-pixel RMS error overall. Such a framework paves the way for accurately tracking surgical instrument keypoints, enabling further downstream use cases. Project and dataset webpage: \url{https://tinyurl.com/mfc-tracker} 
\end{abstract}
\vspace{-10pt}
\section{Introduction}
\label{sec:intro}
\vspace{-5pt}
Robotic minimally invasive surgery (MIS) has seen significant growth over the past two decades~\cite{dupont2021decade}. With more and more robotic surgeries performed every year, there is a need for surgeons to be trained and skilled in robotic surgery. Most of the learning feedback is given through the process of viewing robotic surgery videos by experts, which is time-consuming and subjective. Automated detection and tracking of robotic surgical tools can enable automated surgical skill and learning assessment. Real-time tool tracking can enable computer-assisted interventions, such as marking safe working zones, etc. 

While there have been recent works in surgical instrument detection and instrument segmentation~\cite{iglovikov2018ternausnet,jin2019incorporating}, there have been only a few works focusing on tracking keypoints on surgical tools and instruments, including tool-tips, tool jaw base, and tool shaft point. Estimating keypoints present on surgical tools while performing tasks is a challenging and interesting computer vision task. The ability to track tool-tips in an automated manner from videos can enable downstream video analytics and analysis of surgical tool movement, which can be very useful for surgical skill assessment and judging expertise of motor skill movements~\cite{murali2021velocity}. With these use cases in mind, surgical tool keypoint localization and tracking can become challenging due to changing lighting conditions, temporary occlusions, frame focusing issues, motion blur artifacts, and awkward tool positions and orientations. 

\begin{figure*}[ht!]
    \centering
    \centerline{\includegraphics[width=0.85\linewidth]{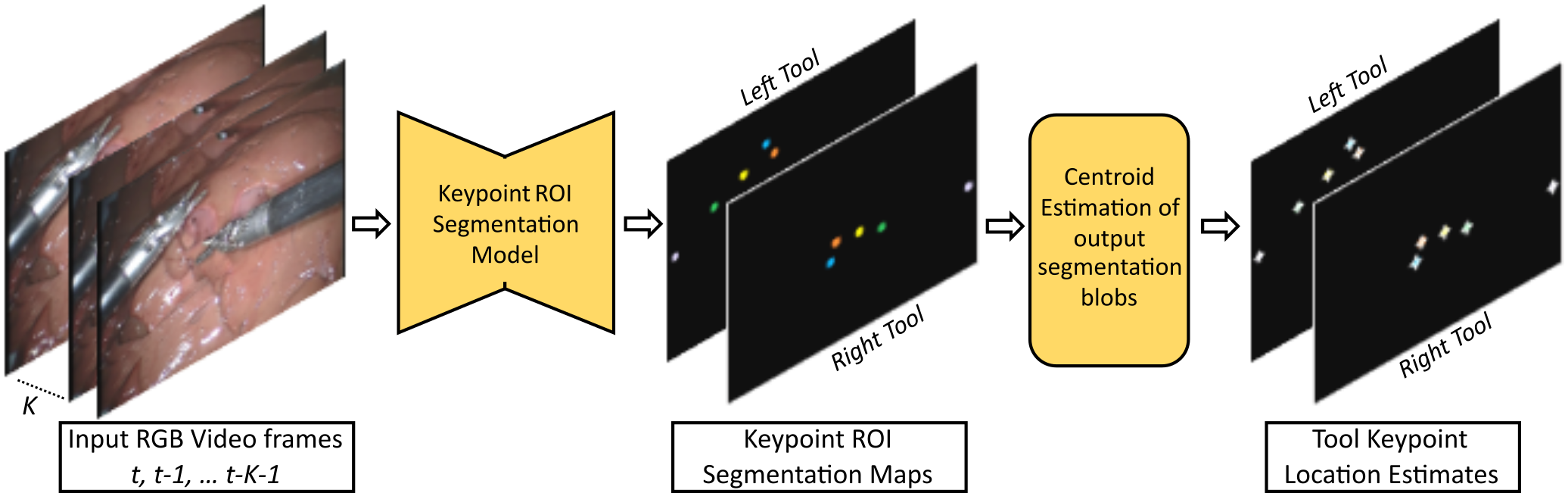}}
    \vspace{-5pt}
    \caption{Proposed keypoint tracking workflow. We perform tracking by (i) segmenting out keypoint regions (keypoint ROIs) using a deep learning-based segmentation model, and (ii) estimating the centroids of segmented output blobs as tool keypoint locations.}
    \vspace{-5pt}
    \label{fig1-teaser}
\end{figure*}

In this work, we propose the use of deep learning-based (DL-based) supervised models that can reliably track surgical tool keypoints such as tool-tips, tool-base, and other relevant keypoints. We pose the tracking problem as a segmentation task to localize small segments (regions) around tool keypoints. By leveraging sophisticated deep learning segmentation architectures~\cite{chen2017rethinking, long2015fully}, and considering optical flow and depth predictions from state-of-the-art optical flow and monocular depth estimation models~\cite{teed2020raft, yang2024depth}, our proposed tracking method (see Fig.~\ref{fig1-teaser}) is able to use a multi-frame context to segment out keypoint regions and subsequently track keypoints of surgical tools from video frames. 

\begin{figure}[h!]
    \centering
    \includegraphics[width=0.95\linewidth]{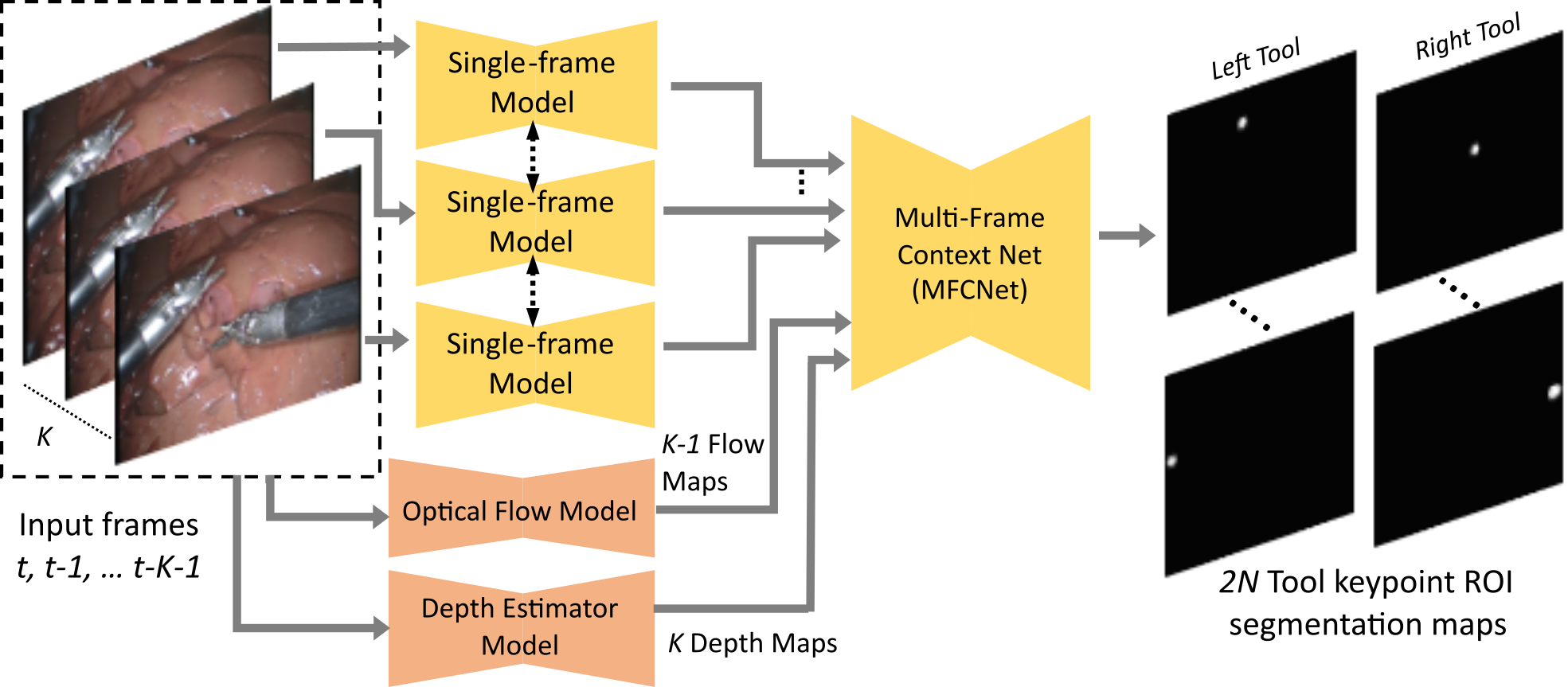}
    \vspace{-5pt}
    \caption{Multi-frame context (MFC) model design for keypoint ROI segmentation. Predictions of $K$ consecutive frames from a trained single-frame context model are computed. Alongside the segmentation maps, $K{-}1$ optical flow maps and $K$ depth maps are also computed. These maps are passed into MFCNet to aggregate multi-frame context and thus estimate accurate keypoint ROI segmentation predictions.}
    \vspace{-15pt}
    \label{fig:mfc-model}
\end{figure}

\vspace{-10pt}
\section{Related Work}
\label{sec:related-work}
\vspace{-5pt}
Surgical instrument detection and segmentation have been of significant interest in recent years, with several works deploying supervised deep learning solutions~\cite{iglovikov2018ternausnet, jin2019incorporating, ni2019rasnet}.

Although significant progress has been made on surgical instrument segmentation, less progress has been made on developing vision-based methods for tracking keypoints of surgical tools. Tracking keypoints and pose are a well-known problem in computer vision. However, most of the modern deep learning solutions focus on estimating human poses and require training on large amounts of data~\cite{wei2016convolutional, zheng2023deep}. 
Non-human keypoint/pose estimation works mainly focus on animal poses~\cite{mathis2018deeplabcut}.
Estimation and tracking for surgical instruments is more challenging, particularly due to scarcity of open-source annotated datasets. Some early works used traditional machine learning-based methods~\cite{sznitman2014fast,allan2015image}, and more recent works~\cite{du2018articulated,colleoni2019deep,du2019patch} have looked at tracking various tool keypoints using convolutional neural networks. Tracking tool-tips and other keypoints in videos can be challenging due to occlusions, motion blur, and video compression artifacts. Our proposed tracking method aims to build an effective tracking method that is capable of handling awkward tool poses and rapid tool motions using only a small annotated dataset. 

\vspace{-10pt}
\section{Methods}
\label{sec:methods}
\vspace{-5pt}
We aim to track various keypoints of surgical tools from videos using novel DL-based trackers. We propose to address the keypoint tracking problem in a two-step manner:
\begin{enumerate}
    \item We first aim to segment out small regions around keypoints (henceforth referred as keypoint regions or keypoint ROIs). This is modeled as a per-pixel multi-class segmentation problem, with each keypoint region of each tool (left/right) being assigned a different class. We aim to train deep learning-based segmentation models that can successfully segment small keypoint regions.  
    \item Once segmentation outputs delineating the keypoint regions are obtained, we localize the keypoints by estimating contiguous segmented regions (blobs) and estimating their centroid as the keypoint coordinate. 
\end{enumerate}
Our workflow is illustrated in Fig.~\ref{fig1-teaser}. 

\vspace{-5pt}
\subsection{Keypoint region segmentation models}
\vspace{-5pt}
In the proposed framework, we group our proposed keypoint region segmentation models into two categories:
\begin{itemize}[leftmargin=*]
    \item \textbf{Single-frame context (SFC) models}. In SFC models, we predict a segmentation output given an input frame at a particular time instance. The segmentation model outputs are independent of past/future video frames. Any deep learning model that performs 2D semantic segmentation~\cite{iglovikov2018ternausnet,chen2017rethinking,long2015fully} given an input image can be thought of as a SFC model. 
    \item \textbf{Multi-frame context (MFC) models}. To make our tracking more robust, we propose a keypoint ROI segmentation model that takes into consideration a multi-frame, moving window context. $K$ consecutive video frames are input, each processed by a single-frame context model (SFC model) to produce keypoint ROI segmentation maps for all $K$ frames. These outputs are refined through an additional refinement model, which we refer to as MultiFrameNet (MFCNet). To provide a better multi-frame context, we propose using optical flow maps and monocular depth estimation maps as auxiliary inputs. Optical flow maps represent the 2D pixel displacement between consecutive video frames, capturing the motion and direction of objects within the scene. Monocular depth estimation maps predict the per-pixel depth from a single RGB frame. Both estimates offer a temporal and geometric context for keypoint positioning. We use standard pretrained models for optical flow and depth estimation (RAFT model~\cite{teed2020raft} and Depth-Anything-V2 model~\cite{yang2024depth} respectively). Thus, we obtain $K{-}1$ optical flow maps (between the current frame and all $K{-}1$ past frames) and $K$ depth maps (from all $K$ frames), and pass their output to MFCNet. This is illustrated in Fig.~\ref{fig:mfc-model}. We propose two different architectures for MFCNet:
    \begin{itemize}[leftmargin=*]
        \item \textbf{MFCNet-Basic (MFCNet-B):} Concatenates all depth maps, optical flow maps, and SFC outputs, then processes them through MFCNet (a 4-layer CNN) for the final keypoint segmentation.  
        \item \textbf{MFCNet-Warp (MFCNet-W):} Warps all depth maps and SFC outputs using the optical flow maps, then processes the warped data through MFCNet (4-layer CNN).   
    \end{itemize}
\end{itemize}

\vspace{-5pt}
\subsection{Keypoint localization}
\vspace{-5pt}
For the tracking of keypoints, we estimate key point coordinates from the output of the segmentation model. The largest segmentation blob is computed from the output segmentation maps using the OpenCV contour estimation function. We predict the centroids of the largest blob as the keypoint location. 
\vspace{-15pt}
\subsection{Dataset Keypoint Annotation and Preparation}
\begin{figure}[h!]
    \centering
    \includegraphics[width=1.0\linewidth]{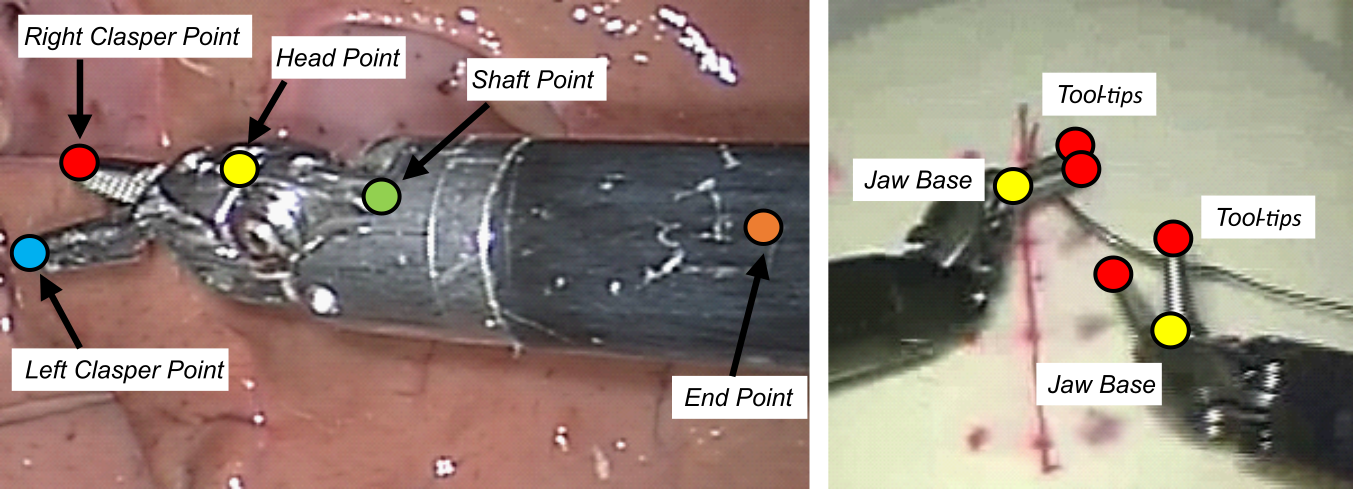}
    \vspace{-15pt}
    \caption{(Left) Keypoint annotations provided in \cite{du2018articulated}. (Right) Self-annotated keypoints on JIGSAWS dataset.}
    \vspace{-15pt}
    \label{fig:endovis15-annotations}
\end{figure}

\vspace{-5pt}
We train and test our models on the dataset given by Du et al. 2018~\cite{du2018articulated}, based on the EndoVis'15 Grand Challenge\footnote{\href{https://endovissub-instrument.grand-challenge.org/EndoVisSub-Instrument/}{https://endovissub-instrument.grand-challenge.org/EndoVisSub-Instrument/}}. This dataset\footnote{\href{https://github.com/surgical-vision/EndoVisPoseAnnotation}{https://github.com/surgical-vision/EndoVisPoseAnnotation}} includes keypoint annotations for robotic MIS tools across 1850 frames (576$\times$720 size), with 5 keypoints per tool (EndPoint, ShaftPoint, HeadPoint, RightClasperPoint, LeftClasperPoint) (see Fig~\ref{fig:endovis15-annotations}). Using these annotations, we generate multi-class ground truth segmentation maps with circular regions (radius $r_d=5$) around each keypoint. Our keypoint ROI segmentation models are trained on a 940-frame training set from 4 videos, resulting in 10 keypoint ROI classes (5 per left/right tool) an additional background class.

We also train and evaluate our tracking framework on the JIGSAWS dataset~\cite{gao2014jhu}, a surgical skill assessment dataset with videos (size 480$\times$640) of 8 participants performing 5 trials each of knot tying, needle passing, and suturing tasks. We annotated 1350 training frames (sampled at 5 FPS) across 6 randomly selected JIGSAWS videos, marking tool tips and tool jaw base points for both left and right tools, using the MATLAB ImageLabeler function. We generated multi-class segmentation maps with small circular regions around each keypoint, similar to our approach with the EndoVis’15 dataset. For testing, 450 frames from 6 additional videos were annotated in the same manner. Our self-annotated dataset can be found at \url{https://tinyurl.com/mfc-tracker}

\vspace{-5pt}
\subsection{Model training and hyperparameter details}
\vspace{-5pt}
Our keypoint region segmentator model generates per-pixel segmentation maps for various keypoints in each frame, assigning each keypoint on the left and right tool a unique class, along with a background class. For the EndoVis 2015 dataset~\cite{du2018articulated}, this results in an 11-class segmentation problem (1 + 5 × 2 = 11). We use a loss function similar to~\cite{iglovikov2018ternausnet}:
\vspace{-3pt}
$$Loss = 0.7\,H - 0.3\,\log J$$
where $H$ is the per-pixel negative log-likelihood (NLL) loss, and $\log J$ represents the logarithm of the Jaccard index across segmented classes. The background class in the NLL loss is weighted down by 1/100 to address class imbalance, and the second term is calculated only for keypoint classes. Single-frame context (SFC) models are trained for 20 epochs using Adam with a learning rate of $3{\times}10^{-5}$ and a batch size of 4. Multi-frame context (MFC) models were trained similarly for 20 epochs with a batch size of 4. We use a pretrained SFC model and finetune it with a lower learning rate of $10^{-6}$, while MFCNet weights are trained with a learning rate of $10^{-4}$ for 20 epochs. We apply a pretrained RAFT model to compute optical flow on a 2x downsampled scale to ease computational load. Across all experiments, the learning rate is decayed by $\gamma{=}0.1$ after 10 epochs. During training, standard data augmentation methods were applied for robust training.

\begin{figure}[h!]
    \centering
    \includegraphics[width=\linewidth]{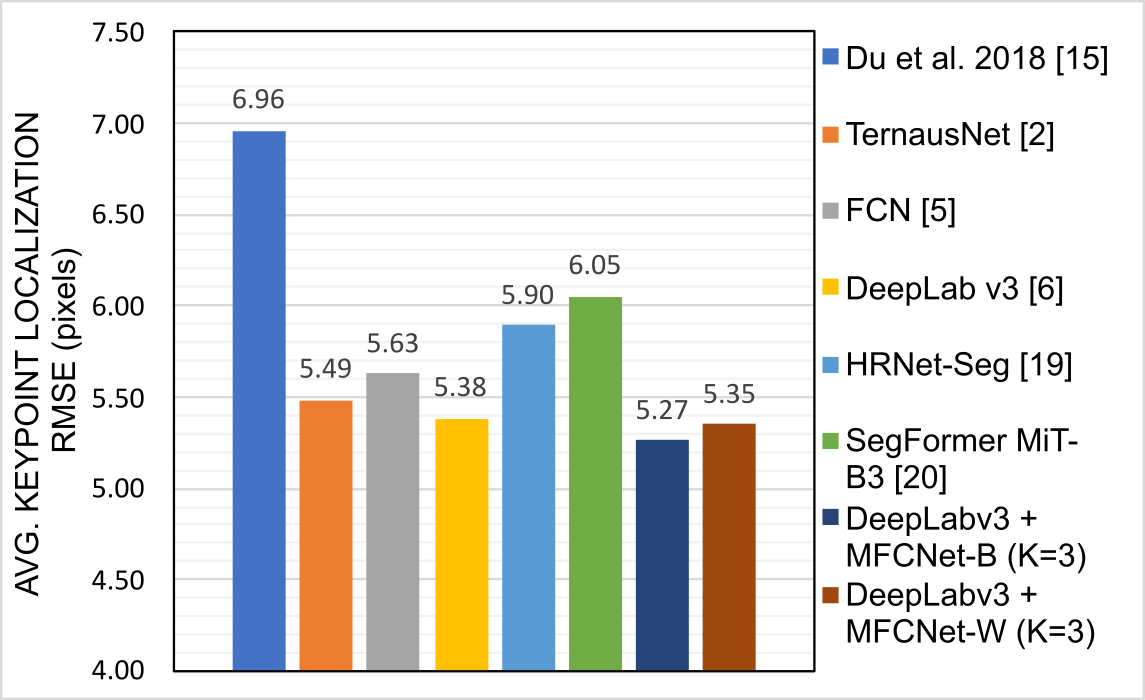}
    \vspace{-15pt}
    \caption{Comparing model performance on the EndoVis 2015 dataset (with Du \textit{et al}. 2018~\cite{du2018articulated} annotations). Showing keypoint localization RMS error (in pixels). Our proposed trackers perform better than \cite{du2018articulated}, and also better than single-frame context models~\cite{iglovikov2018ternausnet,long2015fully,chen2017rethinking, wang2020deep, xie2021segformer}. Values for~\cite{du2018articulated} are taken directly from the corresponding paper.}
    \vspace{-10pt}
    \label{fig:localization-rmse}
\end{figure}

\vspace{-10pt}
\section{Results}
\label{sec:results}
\vspace{-5pt}
\subsection{Results on EndoVis'15 annotated dataset}
\begin{figure}[t!]
    \centering
    \includegraphics[width=\linewidth]{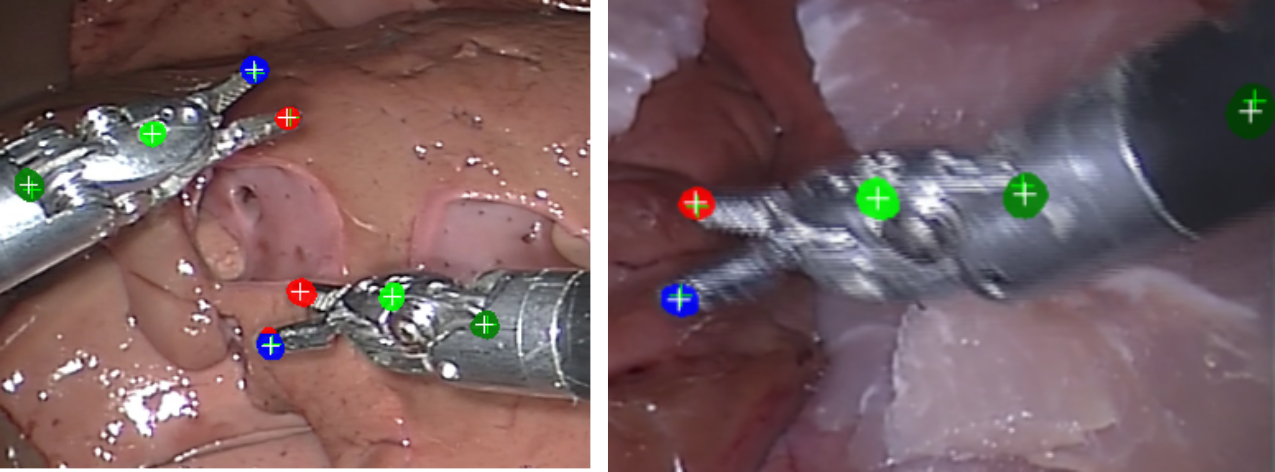}
    \vspace{-15pt}
    \caption{Keypoint localization results - EndoVis'15 dataset. Keypoint ROI segmentations highlighted with colored mask overlays. Green and white crosses indicate GT and estimated keypoint locations.}
    \vspace{-15pt}
    \label{fig:endovis15-res}
\end{figure}

\vspace{-5pt}
We evaluate the trained model on the EndoVis'15 testing dataset (910 frames across 6 videos). The results are shown in Fig~\ref{fig:localization-rmse} and Table~\ref{tab:endovis15-1}. We compare 5 single-frame models i.e. models which predict a segmentation output based on a single input frame only. Compared to the baseline result~\cite{du2018articulated}, the use of better segmentation models such as TernausNet, FCN, DeepLab-v3 gives a lower RMS error (RMSE) even for single-frame context cases. SFC models models such as HRNet~\cite{wang2020deep},  SegFormer~\cite{xie2021segformer} also outperform the baseline but are not the best-performing, likely due to limited training data. We also train both variants of our proposed multi-frame context-driven (MFC) models, using DeepLab-v3 as the SFC base model. Both perform better than the baseline~\cite{du2018articulated} in precision, recall, and localization RMSE, achieving lower RMSE than their SFC counterpart with only a slight drop in precision and recall. While other SFC base models also result in better RMSE when multi-frame context is added, we select DeepLab-v3 for its superior performance as a base model.   
Table~\ref{tab:endovis15-2} shows the ablation experiments performed. The number of frames used in multi-frame model were varied ($K{=}2,3,4$), with the best results occurring with $K{=}3$. With longer windows, there can potentially be a larger drop in quality of flow map estimates, which could result in worse performance for $K{=}4$. Training the multi-frame model (i.e. MFCNet) without depth inputs also results in degraded performance, highlighting the importance of using auxiliary depth inputs for MFCNet. Using other pretrained models such as FlowFormer++~\cite{shi2023flowformer++} (in place of RAFT) yielded similar results. Fig.~\ref{fig:endovis15-res} shows example keypoint tracking results from DeepLabv3+MFCNet-Warp (K{=}3), illustrating the model's accuracy and the ability to handle motion blur.

\begin{table}[h!]
    \vspace{-2pt}
    \centering
    \Footnotesize
    \begin{tabular}{|l|c|c|c|c|c|c|}
        \hline
        \textbf{Model} & \textbf{Avg. Keypoint Prec. / Rec.}         \\ 
        \hline
        Du \textit{et al.} 2018\footnotemark[1]~\cite{du2018articulated} &  83.0 / 83.7 \\ 
        \hline
        DeepLabv3 (single-frame)~\cite{chen2017rethinking} & 100.0 / 91.0        \\ 
        \hline
        DeepLabv3+MFCNet-B (K=3) & 99.9 / 86.6        \\ 
        \hline
        DeepLabv3+MFCNet-W (K=3) & 100.0 / 89.8        \\ 
        \hline
    \end{tabular}
    \vspace{-5pt}
    \caption{\textbf{EndoVis'15 results}. Showing precision and recall scores averaged over 5 keypoints. Our proposed models perform better than Du \textit{et al.} 2018~\cite{du2018articulated}.}
    \vspace{-15pt}
    \label{tab:endovis15-1}
\end{table}
\footnotetext[1]{indicates values taken from results published in~\cite{du2018articulated}.}

\begin{table}[h!]
\vspace{-2pt}
    \centering
    \Footnotesize
    \begin{tabular}{|l|c|}
        \hline
        \textbf{Model} & {\textbf{Localization RMSE}} \\ 
        \hline
        DeepLabv3+MFCNet-W (K=3) & 5.35 $\pm$ 3.67 px \\ 
        \hline
        DeepLabv3+MFCNet-W w/o depth (K=3) & 5.44 $\pm$ 3.75 px \\ 
        \hline
        DeepLabv3+MFCNet-W (K=2) & 5.99 $\pm$ 6.67 px \\ 
        \hline
        DeepLabv3+MFCNet-W (K=4) & 5.40 $\pm$ 4.34 px \\ 
        \hline
    \end{tabular}
    \vspace{-5pt}
    \caption{Ablation experiments on multi-frame models. Showing average and spread of RMS localization errors.}
    \vspace{-5pt}
    \label{tab:endovis15-2}
\end{table}

\begin{figure}[t!]
    \centering
    \includegraphics[width=0.95\linewidth]{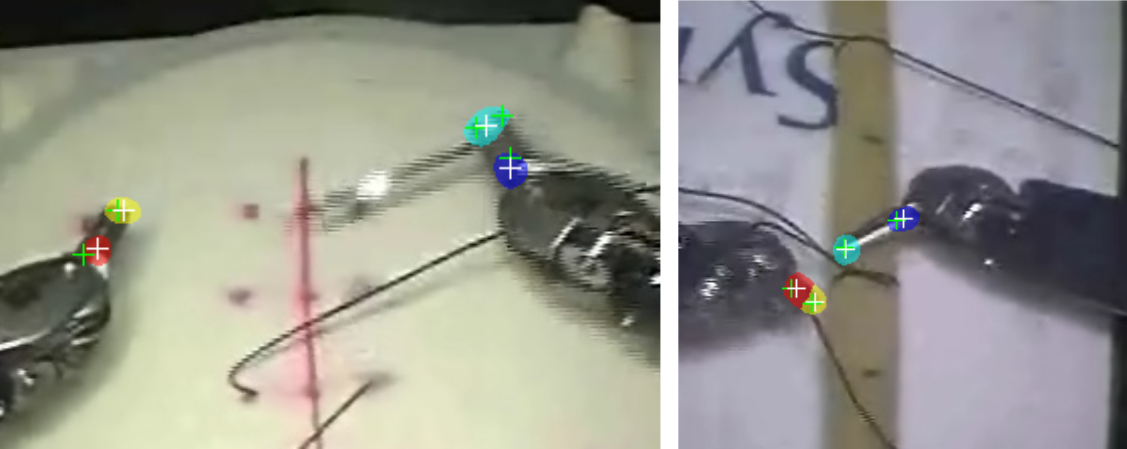}
    \vspace{-5pt}
    \caption{Keypoint localization results - JIGSAWS dataset. Keypoint ROI segmentations highlighted with colored mask overlays. Green and white crosses indicate GT and estimated keypoint locations.}
    \vspace{-15pt}
    \label{fig:jigsaws-res}
\end{figure}

\vspace{-12pt}
\subsection{Results on JIGSAWS dataset}
\vspace{-5pt}
We train and evaluate our model on the JIGSAWS dataset, using our annotations, similar to EndoVis’15. For tooltip detection, however, we identify the two largest blobs and estimate at most two centroids if both are visible. Results are demonstrated in Table~\ref{tab:jigsaws-1}, comparing various single-frame models such as TernausNet16, FCN, DeepLab-v3 and our proposed multi-frame models (DeepLab-v3 + MFCNet-B/W) which take 3 consecutive frames as input. Both proposed multi-frame models achieve the highest (${>}91\%$) keypoint detection accuracy. We observe ${<}4.2$ pixel localization RMSE with the proposed multi-frame models. Fig.~\ref{fig:jigsaws-res} depicts multi-frame model results, showing accurate tracking despite having low resolution image data and handling challenging tool poses. The model inference speeds in terms of frames processed per second (FPS) are shown in Table~\ref{tab:jigsaws-1} (tested on 1 Nvidia A100 GPU). We achieve ${\geq}3.4$ FPS with the MFC models, making them suitable for efficient and fast offline inference. 

\begin{table}[h!]
\vspace{-2pt}
    \centering
    \Footnotesize
    \begin{tabular}{|l|c|c|c|}
        \hline
        \textbf{Model} & {\textbf{Detection}} & {\textbf{Localization }} & \textbf{FPS }\\ 
        & {\textbf{Accuracy}} & {\textbf{RMSE }}& \textbf{}\\
        \hline
        TernausNet16 (single-frame)~\cite{iglovikov2018ternausnet} &  85.4\% & 5.0 $\pm$ 9.3 px & 9.1\\ 
        \hline
        FCN (single-frame)~\cite{long2015fully} &  91.2\% & 4.5 $\pm$ 4.2 px & 8.3\\ 
        \hline
        DeepLabv3 (single-frame)~\cite{chen2017rethinking} & 89.9\% & 4.1 $\pm$ 3.5 px & 7.9\\ 
        \hline
        DeepLabv3+MFCNet-B (K=3)
        & 91.6\% & 4.2 $\pm$ 3.4 px & 3.5\\ 
        \hline
        DeepLabv3+MFCNet-W (K=3)
        & 92.0\% & 4.2 $\pm$ 3.5 px & 3.4\\ 
        \hline
    \end{tabular}
    \vspace{-5pt}
    \caption{\textbf{Keypoint tracking results on JIGSAWS dataset.} Comparison of various models in terms of detection accuracy and localization RMSE for tool keypoints. Inference speeds for each model are also shown.}
    \vspace{-10pt}
    \label{tab:jigsaws-1}
\end{table}

\vspace{-14pt}
\section{Conclusions}
\label{sec:conclusions}
\vspace{-5pt}
We propose deep learning-based surgical tool keypoint tracking models. The tracking is addressed in two stages - keypoint region segmentation, followed by centroid localization. Our multi-frame context-driven models perform accurate keypoint tracking on the EndoVis'15 dataset, performing better than previous baselines~\cite{du2018articulated}, and also better than naive single-frame segmentation models. Our experiments on a more challenging JIGSAWS dataset also yields similar trends. Our proposed framework is versatile - the segmentation model, optical flow model and depth estimator all can be changed to better or more appropriate models if needed. Future work can include generalizing the tracking problem, or training on a larger dataset in a supervised/semi-supervised manner. We envision such tool keypoint tracking models to be used in downstream tasks such as surgical skill/expertise assessment. 

\bibliographystyle{IEEEbib}
\bibliography{strings,refs}

\end{document}